# A Contrast Synthesized Thalamic Nuclei Segmentation Scheme using Convolutional Neural Networks

Lavanya Umapathy[1,2], Mahesh Bharath Keerthivasan[2], Natalie M. Zahr[3], Ali Bilgin[1,2,4] and Manojkumar Saranathan[1,2,4]

[1]Electrical and Computer Engineering, University of Arizona, Tucson, AZ, United States

[2]Medical Imaging, University of Arizona, Tucson, AZ, United States

[3]Psychiatry & Behavioral Sciences, Stanford University, Menlo Park, CA, United States

[4]Biomedical Engineering, University of Arizona, Tucson, AZ, United States

Corresponding Author:

Manojkumar Saranathan, PhD

Associate Professor, Department of Medical Imaging

University of Arizona, Tucson, AZ 85724

Email: manojsar@radiology.arizona.edu

## ABSTRACT

Thalamic nuclei have been implicated in several neurological diseases. White-Matter-nulled Magnetization Prepared Rapid Gradient Echo (WMn-MPRAGE) images have been shown to provide better intra-thalamic nuclear contrast compared to conventional MPRAGE images but the additional acquisition results in increased examination times. In this work, we investigated 3D Convolutional Neural Network (CNN) based techniques for thalamic nuclei parcellation from conventional MPRAGE images. Two 3D CNNs were developed and compared for thalamic nuclei parcellation using MPRAGE images: a) a native contrast segmentation (NCS) and b) a synthesized contrast segmentation (SCS) using WMn-MPRAGE images synthesized from MPRAGE images. We trained the two segmentation frameworks using MPRAGE images (n=35) and thalamic nuclei labels generated on WMn-MPRAGE images using a multi-atlas based parcellation technique. The segmentation accuracy and clinical utility were evaluated on a cohort comprising of healthy subjects and patients with alcohol use disorder (AUD) (n=45). The SCS network yielded higher Dice scores in the Medial geniculate nucleus (P=.003) and Centromedian nucleus (P=.01) with lower volume differences for Ventral anterior (P=.001) and Ventral posterior lateral (P=.01) nuclei when compared to the NCS network. A Bland-Altman analysis revealed tighter limits of agreement with lower coefficient of variation between true volumes and those predicted by the SCS network. The SCS network demonstrated a significant atrophy in Ventral lateral posterior nucleus in AUD patients compared to healthy age-matched controls (P=0.01), agreeing with previous studies on thalamic atrophy in alcoholism, whereas the NCS network showed spurious atrophy of the Ventral posterior lateral nucleus. CNN-based contrast synthesis prior to segmentation can provide fast and accurate thalamic nuclei segmentation from conventional MPRAGE images.

## INTRODUCTION

The thalamus, with its histologically and functionally distinct nuclei, is a deep brain relay organ involved in higher level neurological functions including sleep, arousal, and consciousness. Individual thalamic nuclei have been shown to be differentially involved in multiple neurological diseases including mediodorsal (MD) and pulvinar (Pul) nuclei for schizophrenia (Byne et al. 2001), interlaminar nuclei for Parkinson's Disease (Henderson et al. 2000), limbic and anterior thalamic nuclei for Alzheimer's disease (Braak and Braak 1991; Aggleton et al., n.d.; Low et al. 2019), and anteroventral (AV), pulvinar, and habenular (Hb) nuclei for multiple sclerosis (Planche et al. 2019). Segmentation and analysis of individual thalamic nuclei, in addition to the thalamus, can not only provide more insight in understanding the relationship between thalamic volumes and neurological diseases but also better elucidate the role of thalamic nuclei in healthy and aging brains.

Conventional T1 weighted MRI such as Magnetization Prepared Rapid Gradient Echo (MPRAGE) acquisitions suffer from poor contrast between the thalamus and its surrounding tissues as well as poor intra-thalamic nuclear contrast, making manual or automated thalamic nuclei segmentation challenging. Many automated thalamic parcellation methods reported in the literature are based on diffusion-MRI based techniques (Wiegell et al. 2003; Battistella et al. 2017; Behrens et al. 2003). These images inherently suffer from the poor spatial resolution of the underlying echo-planar imaging acquisition. This along with the isotropic nature of diffusion within the thalamus, a predominantly gray matter region, has limited diffusion-MRI based segmentation to either larger nuclei or nuclei clusters. Recent techniques, such as the Bayesian probabilistic atlas-based thalamus segmentation (Iglesias et al. 2018), available as part of the neuroimaging package Freesurfer (https://surfer.nmr.mgh.harvard.edu), use T1 MPRAGE acquisitions. However, the processing time for Freesurfer-based segmentation is in the order of several hours.

Recently, a white-matter-nulled variant of the MPRAGE sequence, hereafter referred to as WMn-MPRAGE, was demonstrated to provide excellent white matter-thalamic contrast to clearly identify thalamus boundaries for segmentation (Tourdias et al. 2014), and maximize intra-thalamic nuclear contrast at 7T (Saranathan et al. 2015). A multi-atlas segmentation technique using WMn-MPRAGE imaging called THOMAS (Thalamus Optimized Multi Atlas Segmentation) (Su et al. 2019) has also been proposed. THOMAS uses 20 anatomical priors with manual delineations by a radiologist, guided by the Morel atlas, to segment the thalamus into 11 nuclei and the mammillothalamic tract in 15 minutes or less. Evaluated on healthy volunteers and patients with MS by comparing against manual annotations, THOMAS achieved high Dice scores of at least 0.85 for the larger nuclei and at least 0.7 for smaller structures. It also performed significantly better than Freesurfer-based segmentation, especially for small nuclei such as lateral and medial geniculate nucleus (Su et al. 2019).

Although WMn-MPRAGE images provide excellent intra-thalamic nuclei contrast, routine clinical studies acquire only conventional (i.e. CSF-nulled) MPRAGE images. The inclusion of a specialized WMn-MPRAGE sequence will add 7-10 minutes of scan time to the acquisition protocol. Moreover, numerous data repositories like Alzheimer's Disease Neuroimaging Initiative (ADNI) or Open Access Series of Imaging Studies (OASIS) only have standard MPRAGE data.

Segmentation techniques such as Freesurfer and THOMAS rely on success of the nonlinear registration process to accurately transfer thalamic nuclei labels from atlas space to native space. Convolutional Neural Networks (CNNs) can provide a fast, accurate, and registration-free alternative for thalamus segmentation. CNNs can also be used to non-linearly map from CSF-nulled MPRAGE contrast to WM-nulled MPRAGE with improved intra-thalamic nuclei contrast. In this work, we propose the use of a deep learning-based 3D CNN framework for segmenting thalamic nuclei from conventional MPRAGE images. We first use a multi-scale 3D CNN scheme to segment thalamic nuclei directly from MPRAGE data. We refer to this as native contrast segmentation (NCS) scheme. We hypothesize that, synthesizing WMn-MPRAGE images from MPRAGE images can improve thalamic nuclear conspicuity and, in turn, segmentation accuracy. To test this hypothesis, we propose the introduction of a multi-scale 3D CNN scheme for WMn-MPRAGE contrast synthesis prior to the CNN framework for segmentation. We refer to this as synthesized contrast segmentation (SCS) scheme. We compare the performance of the thalamic nuclei segmentation CNN trained directly on MPRAGE images (i.e. NCS) with those using synthesized WMn-MPRAGE images (i.e. SCS) on a cohort of healthy subjects (n=12). Finally, we use the CNNs proposed in this work on a cohort of patients (n=45) with alcohol use disorder (AUD) and age-matched control to study the effect of AUD on thalamic nuclear atrophy.

## MATERIALS AND METHODS

### Study Cohort

For generalizability, images acquired on two different scanners were pooled for training. This comprised of data from 18 healthy individuals acquired on a 3T MAGNETOM Prisma (Siemens Medical Solutions USA, Inc., Malvern, PA, USA) and 17 healthy individuals acquired on a 3T Signa (General Electric Healthcare, Waukesha, WI) MR scanner, each subject being scanned with both conventional MPRAGE and WMn-MPRAGE protocols. Of these 35 individuals, 33 subjects were retained for training, 2 were used for validation. Note that the training and validation cohorts consisted entirely of healthy individuals with no known pathologies. Two independent test cohorts were used to demonstrate generalizability. The first cohort, referred to as test cohort 1, consisted of 12 healthy subjects. A second test cohort (n=45) consisting of 22 subjects with AUD and 23 gender- and age- matched healthy controls, was used to demonstrate clinical utility of the proposed technique. We refer to this cohort as the AUD cohort. All imaging data was acquired after prior informed consent and adhering to institutional review board guidelines.

### CNN Architecture Overview

Figure 1 shows an overview of the two proposed thalamic nuclei segmentation frameworks. Fig. 1a shows the 3D multi-scale Native Contrast Segmentation (NCS) CNN framework that predicts the whole thalamus and individual thalamic nuclei on MPRAGE images. The THOMAS multi-atlas label fusion based algorithm described previously (Su et al. 2019) was used to generate ground truth labels for the thalamus and 11 thalamic nuclei using corresponding co-registered WMn-MRAGE images. Fig. 1b shows an alternative Synthesized Contrast Segmentation (SCS) CNN scheme where a 3D multi-scale contrast synthesis CNN is first used to synthesize WMn-MPRAGE images from MPRAGE images. These synthesized images are then used as inputs to another segmentation network.

The multi-scale thalamic nuclei segmentation CNN has a UNET (Ronneberger, Fischer, and Brox 2015) like encoder-decoder framework as shown in Fig. 2a. Multi-resolution feature maps are first used to predict thalamus label. These feature maps, concatenated with the predicted thalamus label, are then used to localize individual thalamic nuclei. Dice loss was used for the thalamus and a multi-label Dice loss was used for thalamic nuclei. If $G_i$ and $P_i$ refer to the ground truth annotations and predictions for the i$^{th}$ label, respectively, the multi-label Dice loss used in this work is defined as follows:

$$L_{multi-label\ Dice} = -\sum_{i}^{C} \frac{|G_i \cap P_i|}{|G_i| + |P_i|}$$

Here, C refers to the total number of nuclei labels. The total segmentation loss is the sum of errors in the prediction of the thalamus using Dice loss and its individual nuclei using multi-label Dice loss.

The architecture for the multi-scale WMn-MPRAGE synthesis CNN also uses an encoder-decoder architecture similar to the segmentation CNN (Fig. 2b) except for the loss formulation. Along with an intensity similarity-based loss function, we also use the perceptual loss using a VGG16 model (Simonyan and Zisserman 2015) pre-trained on ImageNet, similar to some other works on contrast synthesis in MR images (Dar et al. 2019). If $W$ and $W_{syn}$ are the ground truth WMn-MPRAGE image and the WMn-MPRAGE image synthesized from the corresponding MPRAGE image, respectively, the total synthesis loss function is given as:

$$L\{W, W_{syn}\} = \left\lVert W - W_{syn} \right\rVert_1 + \left\lVert F(W) - F(W_{syn}) \right\rVert_2$$

The first term represents the intensity similarity-based loss defined as the $L_1$ norm-error between the ground truth and the synthesized images. The second term, corresponding to the $L_2$ norm error of the perceptual loss, ensures visual similarity of the synthesized images and the reference images. The function F(.) extracts feature maps from the third rectified linear unit layer of the pre-trained VGG16 model to compare features from ground truth and synthesized images.

**CNN Data Generation**

Whole thalamus and thalamic nuclei labels were generated from WMn-MPRAGE images using THOMAS. The isotropic resolution MPRAGE images and the corresponding WMn-MPRAGE images were co-registered using an affine transformation to train all segmentation CNNs. The co-registered images were brain extracted using FSL's brain extraction toolbox (Smith 2002), N4-bias corrected (Tustison et al. 2010), and intensity normalized using histogram-based contrast stretching to scale intensities in the range of [0, 1]. Contrast stretching was performed to clip intensities below the 1st percentile and above the 99th percentile to avoid outlier-based clipping that may affect traditional min-max based intensity normalization.

The segmentation CNNs were trained using an image-based scheme. From each subject, training data for the segmentation CNN were generated by cropping and extracting 2.5D (192x192x5) images using a sliding window. The WMn-MPRAGE synthesis network was trained using a patch-based scheme. The co-registered image pairs (MPRAGE and WMn-MPRAGE images) were used to extract overlapping 2.5D patches (64x64x5) using a sliding window over the entire brain. The patches were extracted only from regions within the brain mask for each subject. Augmented images were generated in run time on every training image batch. Data augmentation options for the segmentation CNN training included scaling and shearing transformations. Additionally, random in-plane and through-plane rotations were also used for training of the patch-based synthesis CNN.

**CNN Experiments and Implementation**

To understand how thalamic nuclei segmentation performance changes with loss function, we also trained a NCS CNN with weighted categorical cross entropy loss (WCCE). Here, a binary cross-entropy loss was used for thalamus and WCCE was used for thalamic nuclei. The WCCE loss function is defined as:

$$L_{wcce} = -\sum_n \sum_i^C w_i \, g_n \log p_n$$

Here, C is the number of labels, $w_i$ is the weight associated with the i[th] label, $g_n$ and $p_n$ refer to the nth pixel of the ground truth (G) and prediction (P), respectively.

The training parameters for the multi-scale thalamic nuclei segmentation CNN are: loss function for thalamus = Dice, loss function for thalamic nuclei = multi-label Dice, epochs = 50, batch size = 10, optimizer = Adam, learning rate = 0.001, and decay = 0.1. The training parameters for the multi-scale WMn contrast synthesis CNN are: Loss function for image similarity = L1, perceptual loss = L2, epochs = 50, batch size = 10, optimizer = Adam, learning rate = 0.001, decay = 0.1.

**Evaluation of CNN Performance**

The whole thalamus and individual thalamic nuclei labels were predicted on MPRAGE images using the two segmentation CNNs. Performance of the segmentation CNN was evaluated using Dice overlap score and volume difference (VD). If G and P refer to the ground truth THOMAS labels and predictions, respectively, then, Dice = $\frac{2|G \cap P|}{|G|+|P|}$ and VD = $\frac{|V_G - V_P|}{V_G} \times 100$. Here, $V_G$ corresponds to the volume of mask $G$. The performance of the multi-scale WMn contrast synthesis CNN was evaluated using root mean squared error (RMSE), structural similarity (SSIM), and peak signal-to-noise (PSNR) ratio.

**Statistical Analysis**

A recent work by Zahr et al (2020) used WMn-MPRAGE images and THOMAS to report a significant atrophy in the Ventral lateral posterior (VLp) nucleus on a cohort of patients with alcohol use disorder (AUD) and age-matched controls. We used the AUD cohort from this work to characterize atrophy in thalamic nuclei volumes in alcoholism, but using MPRAGE images instead of WMn-MPRAGE.

The 11 thalamic nuclei were combined into four groups similar to Zahr et a al (2020) : anterior, lateral, posterior, and medial. The nuclei breakdown for these groups is shown in Table 1. ANCOVA analysis was first performed on the predicted volumes from the NCS and SCS CNNs to understand the relationship between volumes of the nuclei groups and diagnosis (controls vs alcoholics), after accounting for age and intercranial volume (ICV) estimated using Freesurfer segmentation (Buckner et al. 2004). If a significant relationship was observed within any of these groups, ANCOVA analysis was repeated for individual nuclei within that group to further understand relationship between thalamic nuclei volumes and diagnosis.

A Bland Altman analysis was performed to assess agreement between ground truth and volumes predicted by NCS and SCS CNNs. Pearson's correlation coefficient, coefficient of variation (CV) and percent reproducibility coefficient (RPC) statistics were also computed. A paired two-sided t-test ($\alpha = 0.05$) was used to compare if there were any significant differences in the segmentation performance of synthesized WMn-MPRAGE images compared to MPRAGE images.

## RESULTS

The end-to-end prediction time for a pre-processed MPRAGE volume (~ 256x256x210) was approximately 12 seconds, with 4 seconds for synthesizing WMn-MPRAGE images and 8 seconds for segmenting the thalamus and its nuclei.

Table 2 compares the changes in segmentation accuracy based on the choice of loss function on test cohort 1 (n=12 healthy subjects). Although the segmentation CNN trained using WCCE loss function had a higher Dice score for larger nuclei such as Pul, MD, and VLp, it had significantly higher VD in smaller nuclei such as LGN and MGN (P<.008). For the rest of the results presented in subsequent sections, we show NCS and SCS CNNs trained using multi-label Dice loss function.

### Synthesized WMn-MPRAGE images

Synthesized WMn-MPRAGE images generated from the contrast synthesis CNN are compared to input MPRAGE and target WMn-MPRAGE images in Figure 3. We see that the synthesized WMn-MPRAGE images are able to accurately mimic the contrast between thalamus and white matter similar to the WMn-MPRAGE images. A quantitative comparison of the synthesized images on both test cohorts is presented in Table 3. The synthesized WMn-MPRAGE images have a high structural similarity (93.7% and 92.1%) and PSNR (23.5 and 21.3) on both test cohorts.

### Effect of Contrast Synthesis on segmentation performance

A qualitative comparison of the thalamic nuclei segmentation predictions from NCS and SCS CNNs for a test subject are shown in Figure 4. The ground truth THOMAS labels for the left hemisphere are overlaid as outlines for each label. Note the improved intra-thalamic contrast as well as thalamus-WM boundary in SCS (arrows) compared to NCS. Both the CNN models are able to accurately identify thalamic nuclei relative to the ground truth for the larger nuclei. However, the SCS CNN segments smaller nuclei such as VA, Hb, and MTT better than the NCS CNN. Table 4 presents a quantitative comparison of segmentation performance of the two segmentation CNNs on the AUD cohort. We observed that synthesizing WMn-MPRAGE images prior to segmentation significantly reduced overall VD of the whole thalamus (P=.008). For individual nuclei, the predictions from SCS CNN showed a significantly higher Dice for smaller nuclei such as MGN (P=.003) and CM (P=.01) and significantly lower VD for VA (P=.001) and VPl (P=.01). Although the Dice scores were relatively comparable, SCS CNN made smaller errors in volume estimates compared to NCS in most of the nuclei.

Figure 5 shows thalamic nuclei predictions for the nuclei in the four nuclei groups (anterior, ventral, posterior, and medial) from SCS CNN overlaid on multi-planar views of synthesized WMn-MPRAGE images. The reference THOMAS labels are outlined on the left thalamus. The coronal view shows the LGN and MGN nuclei. The predicted labels for the smaller nuclei such as AV, VA, VLa, CM, LGN, and MGN agree well with the THOMAS outlines.

**Effect of diagnosis on thalamic nuclei volumes in AUD cohort**

The least-squares mean (and standard error) from the ANCOVA analysis on the AUD cohort are presented in Table 5. There was a significant effect of diagnosis on lateral nuclei group volumes after controlling for age and ICV. This reduction in lateral group of nuclei volumes in alcoholics was observed in volumes obtained from both SCS (F [1, 39] = 6.96, $p$ = 0.01) and NCS CNNs (F [1, 39] = 6.92, $p$ = 0.01). We did not observe any significant effect of diagnosis on other nuclei groups (anterior, medial, or posterior). Among the constituent nuclei of the lateral group (VLp, VPl, VA, VLa), both SCS (F [1, 39] = 6.68, $p$ = 0.01) and NCS CNNs (F [1, 39] = 5.96, $p$ = 0.04) segmentations showed a significant atrophy in VLp nucleus volumes in alcoholism. Note that these results are consistent with the previously reported study by Zahr et al (2020) where a significant atrophy was observed in the VLp nucleus in alcoholics but using WMn-MPRAGE images segmented using THOMAS.

In addition to VLp, we noticed that the NCS CNN also showed a significant reduction in VPl (F [1, 39] = 7.92, p = 0.008) nucleus for subjects with AUD. Note that Zahr et al (2020) or the SCS scheme did not show significant atrophy in VPl nucleus. We generated Bland-Altman plots to assess agreement in CNN-predicted volumes and the ground truth for the whole thalamus, VLp, and the VPl nuclei for the AUD cohort (Figure 6). The plots also color-code the data points with the controls shown in blue and the AUD subjects in red. We can see a stronger correlation between ground truth and volume estimates from SCS CNN compared to NCS CNN. The limits of agreement were also tighter when using synthesized WMn-MPRAGE images for the whole thalamus (Fig. 6d) and VPl nucleus (Fig. 6f) compared to using MPRAGE images (Fig. 6a and Fig. 6c).

Bland-Altman plots for NCS CNN show a higher CV for the thalamus (7% vs 4% from SCS CNN) and VPl nucleus (17% vs 10% from SCS CNN). This is also consistent with the significantly higher VD in the thalamus (4.9% in NCS vs 2.9% from SCS, P< .008) and VPl (13.3% in NCS vs 8.1% from SCS, P< .008) observed in Table 4.

Figure 7 illustrates an example AUD test subject where the NCS CNN fails to predict individual thalamic nuclei in the presence of thalamic lesions. The predicted thalamic nuclei from NCS and SCS CNN are overlaid on the MPRAGE and the synthesized WMn-MPRAGE images, respectively, for comparison. We see that the thalamic lesions appear brighter in the synthesized WMn-MPRAGE contrast. Although both the CNN frameworks are trained on healthy individuals, synthesizing WMn-MPRAGE contrast (SCS CNN) is able to better delineate thalamic nuclei boundaries, especially of Pul, CM, and VPl nuclei, even in the presence of thalamic lesions.

**DISCUSSION**

In this work, we have proposed novel 3D CNN based thalamic nuclei segmentation frameworks for conventional MPRAGE images. We developed and tested two different approaches: (a) segmentation CNN trained on native CSF-nulled MPRAGE images (NCS) and (b) segmentation on synthesized WMn-MPRAGE images MPRAGE (SCS) by introducing a contrast synthesis

CNN prior to segmentation. These were validated on two different cohorts- one containing healthy subjects and one consisting of a mix of healthy controls and subjects with AUD. For individual thalamic nuclei, the synthesized WMn-MPRAGE yielded significantly higher Dice and lower volume difference for some nuclei compared to conventional MPRAGE. The use of synthesized WMn-MPRAGE images resulted in significantly lower error in the prediction of total thalamus volumes and accurate prediction of the thalamic nuclear atrophy in AUD patients

**Clinical Utility in AUD cohort**

Segmentation results indicate significant improvements in the Dice metric with the SCS for MGN and CM nuclei as well as significantly lower volume difference in VA and VPl nuclei compared to NCS. This is further corroborated by the improved agreement in predicted volumes and ground truth labels as shown in the Bland-Altman analysis. More significantly, we observed that NCS under-estimated volumes from subjects with AUD, artificially inflating the group differences in subjects with thalamic atrophy due to the underlying pathology. Furthermore, the volume difference for the VPl nucleus with respect to the ground truth was 13.3% ± 12.7% compared to only 8.1% ± 6.2% from the SCS. Together, these explain the spurious atrophy in VPl nucleus volumes observed in alcoholics when using volumes computed from the direct trained segmentation network (i.e. NCS). Using the synthesized WMn-MPRAGE images, we demonstrated atrophy only in VLp nucleus, as previously reported by Zahr et al (2020) using THOMAS and true WMn-MPRAGE data, attesting to the sensitivity, accuracy, and potential utility of the proposed technique.

**Computational Efficiency**

Manual annotations for the thalamus and thalamic nuclei on structural MR images, although considered gold-standard, are tedious; often requiring hours to accurately delineate. Automated thalamic nuclear segmentation techniques on MRI provide a promising alternative to time-consuming manual annotations. Atlas-based techniques rely on registration to transfer thalamic nuclei labels from an atlas space to native structural MR space. In addition to being computationally burdensome, the accuracy of the segmentation is often limited by the success of the registration process. Deep learning-based segmentation techniques can provide a fully automated, fast, reliable, and a registration-free alternative to atlas-based thalamic segmentations. The CNN-based segmentation framework provides accurate and fast thalamic parcellations (11 nuclei and MTT) with prediction times in the order of seconds for a 3D volume. This is a significant reduction compared to multi-atlas schemes (15 min) or probabilistic schemes like Freesurfer (several hours).

**Comparison to existing literature**

A majority of the work on thalamic parcellations are based on diffusion-MRI based techniques (Battistella et al. 2017; Wiegell et al. 2003). Recently, Iglehart et al (2020) published a systematic comparison of thalamic nuclei segmentation techniques based on structural, diffusion, and resting state functional MR acquisitions. In this work, the authors showed that parcellations on structural MR images were more accurate in the delineation of smaller thalamic structures such as AV, Hb, and MGN compared to Diffusion Tensor Imaging (DTI).

Recently, a multi-planar 2D CNN-based method was proposed for thalamic nuclei segmentation on WMn-MPRAGE images, where Majdi et al (2020) demonstrated preliminary results of thalamic nuclei segmentation from conventional MPRAGE images using joint training from WMn-MPRAGE and MPRAGE contrasts. A cascade of 2D CNNs first predicted the thalamus and then used a cropped bounding box for the thalamus to subsequently predict individual thalamic nuclei. They showed that this 2D CNN framework also significantly outperformed Freesurfer-based segmentation in terms of Dice scores, where the Freesurfer-based segmentation yielded significantly lower Dice scores for small nuclei such as AV, VLa, CM as well as LGN and MGN (Majdi et al. 2020).

It should be noted that our proposed 3D NCS scheme to directly predict thalamic nuclei segmentation on MPRAGE images is conceptually similar to the multi-planar network of Majdi et al (2020) used for MPRAGE contrast-based segmentation except that we use a single network to generate the thalamus and individual nuclei as opposed to a cascaded approach. Our results show that when acquisition of WMn-MPRAGE images is not possible, synthesizing WMn-MPRAGE contrast images (i.e. SCS) prior to segmentation can improve accuracy compared to direct segmentation on MPRAGE images (NCS). The segmentation on synthesized contrast yielded improved Dice scores and reduced volume difference for smaller nuclei such as MGN, CM, and VA compared to segmentation on native MPRAGE contrast, with ability to predict atrophy in a disease cohort. It should be pointed out that the focus of our work was on validating our initial hypothesis that synthesizing WMn-MPRAGE images from MPRAGE images can improve segmentation accuracy, rather than identifying the most novel architecture for thalamic nuclei segmentation on MPRAGE images.

**Limitations and Future work**

A limitation of our work is the lack of manual annotations to train the segmentation CNNs. Manual annotations by experts, although considered gold standard, can be impractical to obtain for a large cohort such as ours. The choice of target labels for the thalamus and individual thalamic nuclei, especially one generated from another automated segmentation technique, may affect the segmentation accuracy. However, the multi-atlas framework based thalamic nuclei labels generated from THOMAS have previously been validated against manual annotations (Su et al. 2019). Evaluated on healthy volunteers and patients with MS by comparing against manual annotations, THOMAS achieved high Dice scores of at least 0.85 for the larger nuclei and at least 0.7 for smaller structures. Su et al (2019) compared structural-MRI based THOMAS and Freesurfer-based thalamic nuclei segmentation to manual annotations on a small subset of cases and found that THOMAS outperformed Freesurfer in terms of Dice for smaller nuclei such as CM, AV, and VA as well as in larger nuclei such as VLp and Pul. With improved performance against benchmarks, THOMAS-generated labels act as excellent surrogates for time-consuming manual annotations.

In this study, the segmentation network was trained on a normal cohort and evaluated in subjects with AUD. The clinical applicability of the proposed technique to other neurological diseases that differentially affect thalamic nuclei volumes as well in the presence of lesions or metal electrode artifacts (for e.g. from deep brain stimulation leads) needs further investigation. Online repositories such as ADNI or OASIS with conventional MPRAGE images can be used to explore the relationship between thalamic nuclei volumes and cognition. A fast and reliable MPRAGE-based thalamic parcellation technique such as ours would allow retrospective analysis of these previously acquired neuroimaging data.

**CONCLUSION**

In this work, we propose the use of a cascade of multi-scale 3D CNNs to synthesize WM-nulled contrast from conventional MPRAGE images, prior to thalamic nuclei segmentation. We were able to confirm our hypothesis that synthesizing WMn-MPRAGE images from routine clinical MPRAGE images prior to segmentation can improve the performance of the segmentation CNN. We were able to reproduce the same atrophy in VLp nucleus observed with WMn-MPRAGE images with THOMAS segmentation, attesting to its sensitivity, accuracy, as well as clinical utility.

**TABLES**

Table 1: Nuclei group breakdown into constituent nuclei

| Groups | Nuclei |
|---|---|
| Anterior | Anterior Ventral nucleus (AV) |
| Lateral | Ventral lateral posterior nucleus (VLp) |
| | Ventral lateral anterior nucleus (VLa) |
| | Ventral anterior nucleus (VA) |
| | Ventral posterior lateral nucleus (VPl) |
| Posterior | Pulvinar nucleus (Pul) |
| | Medial geniculate nucleus (MGN) |
| | Lateral geniculate nucleus (LGN) |
| Medial | Mediodorsal nucleus (MD) |
| | Centromedian nucleus (CM) |
| | Habenular nucleus (Hb) |
| Others | Mammillothalamic tract (MTT) |

Table 2: Effect of CNN loss function on segmentation accuracy in test cohort 1 (n=12).

| | | Dice Scores | | Volume Difference | |
|---|---|---|---|---|---|
| | | WCCE[a] | Multi-label Dice | WCCE | Multi-label Dice |
| | **Thalamus** | 0.907 ± 0.03 | 0.915 ± 0.03 | 7.7 ± 7.3 | 6.1 ± 7.1 |
| Lateral | **VLp** | **0.842 ± 0.03*** | 0.82 ± 0.04 | 6.4 ± 5.2 | 7.3 ± 7.4 |
| | **VPl** | 0.74 ± 0.06 | 0.66 ± 0.12 | 16.9 ± 12.4 | 17.2 ± 18.4 |
| | **VA** | 0.723 ± 0.08 | 0.72 ± 0.10 | 20.3 ± 10.0 | 14.0 ± 13.8 |
| | **VLa** | 0.638 ± 0.11 | 0.62 ± 0.11 | 45.2 ± 39.5 | 23.8 ± 19.9 |
| Posterior | **Pul** | **0.888 ± 0.02*** | 0.84 ± 0.05 | **4.7 ± 3.9*** | 10.9 ± 5.4 |
| | **MGN** | 0.72 ± 0.08 | **0.77 ± 0.07*** | 27.0 ± 13.3 | **9.3 ± 8.0*** |
| | **LGN** | 0.63 ± 0.20 | **0.74 ± 0.14*** | 29.4 ± 20.4 | **13.2 ± 10.6*** |
| Medial | **MD** | **0.86 ± 0.07*** | 0.80 ± 0.12 | 12.5 ± 13.6 | 9.0 ± 10.2 |
| | **CM** | 0.67 ± 0.06 | 0.57 ± 0.17 | 52.1 ± 29.2 | 37.6 ± 26.3 |
| | **Hb** | 0.26 ± 0.22 | **0.56 ± 0.28*** | 48.6 ± 33.3 | 25.4 ± 32.8 |
| | **AV** | 0.699 ± 0.17 | 0.66 ± 0.14 | 40.2 ± 24.3 | 24.0 ± 21.1 |

[a]WCCE = Weighted Categorical Cross Entropy

*P<0.008, two-sided paired t-test

Table 3: Evaluation (mean ± SD) of synthesized WMn-MPRAGE images

| | Test Cohort 1 | AUD Cohort |
|---|---|---|
| Root mean squared error (RMSE) | 0.067 ± 0.007 | 0.086 ± 0.009 |
| Peak signal to noise (PSNR) | 23.5 ± 0.968 | 21.3 ± 0.829 |
| Structural similarity (SSIM) | 0.937 ± 0.010 | 0.921 ± 0.017 |

Table 4: Comparison of segmentation performance (Mean ± SD) on Alcohol Use Disorder Cohort (n=45)

|  | Dice Score | | Volume Difference | |
| --- | --- | --- | --- | --- |
|  | NCS[a] | SCS[b] | NCS | SCS |
| Thalamus | 0.92 ± 0.03 | 0.93 ± 0.01 | 4.9 ± 5.0 | **2.9 ± 2.3*** |
| VLp | 0.84 ± 0.04 | 0.84 ± 0.04 | 6.4 ± 5.0 | 7.1 ± 4.7 |
| VPl | 0.73 ± 0.12 | 0.73 ± 0.1 | 13.3 ± 12.7 | **8.1 ± 6.2*** |
| VA | 0.77 ± 0.04 | 0.78 ± 0.05 | 10.8 ± 8.6 | **7.3 ± 6.5**** |
| VLa | 0.65 ± 0.12 | 0.67 ± 0.1 | 20.1 ± 16.0 | 16.8 ± 12.1 |
| Pul | 0.86 ± 0.09 | 0.88 ± 0.03 | 8.1 ± 12.2 | 5.4 ± 4.6 |
| MGN | 0.76 ± 0.08 | **0.78 ± 0.07**** | 11.6 ± 9.7 | 9.6 ± 7.1 |
| LGN | 0.76 ± 0.09 | 0.77 ± 0.08 | 15.1 ± 11.6 | 13.5 ± 13.4 |
| MD | 0.86 ± 0.05 | 0.87 ± 0.03 | 9.0 ± 10.2 | 6.7 ± 4.6 |
| CM | 0.72 ± 0.12 | **0.76 ± 0.07*** | 14.6 ± 14.5 | 12.3 ± 11.2 |
| Hb | 0.68 ± 0.11 | 0.69 ± 0.08 | 17.2 ± 12.8 | 14.4 ± 7.9 |
| AV | 0.75 ± 0.11 | 0.74 ± 0.09 | 26.9 ± 51.7 | 24.9 ± 59.8 |

*P=.01, two-sided paired t-test

**P<.005, two-sided paired t-test

[a]NCS: Native Contrast Segmentation CNN

[b]SCS: Synthesized Contrast Segmentation CNN

Table 5: Thalamic atrophy in alcoholism (ANCOVA[a])

|  | Lateral Group | | |
| --- | --- | --- | --- |
|  | Controls | AUD | P-value |
| NCS | 3133.1 ± 69.7 | 2865.2 ± 71.5 | 0.01* |
| SCS | 3153.3 ± 49.0 | 2964 ± 50.2 | 0.01* |
|  | **Ventral lateral posterior** | | |
|  | Controls | AUD | P-value |
| NCS | 1787.6 ± 34.9 | 1678.5 ± 35.8 | 0.04* |
| SCS | 1850.2 ± 31 | 1733.2 ± 31.7 | 0.01* |
|  | **Ventral posterior lateral** | | |
|  | Controls | AUD | P-value |
| NCS | 604.1 ± **23.4** | 507.8 ± **24** | 0.008* |
| SCS | 567.8 ± 12.8 | 537.1 ± 13.1 | 0.1 |

[a]Least squares mean ± standard error from the ANCOVA analysis (n=45).

# FIGURES

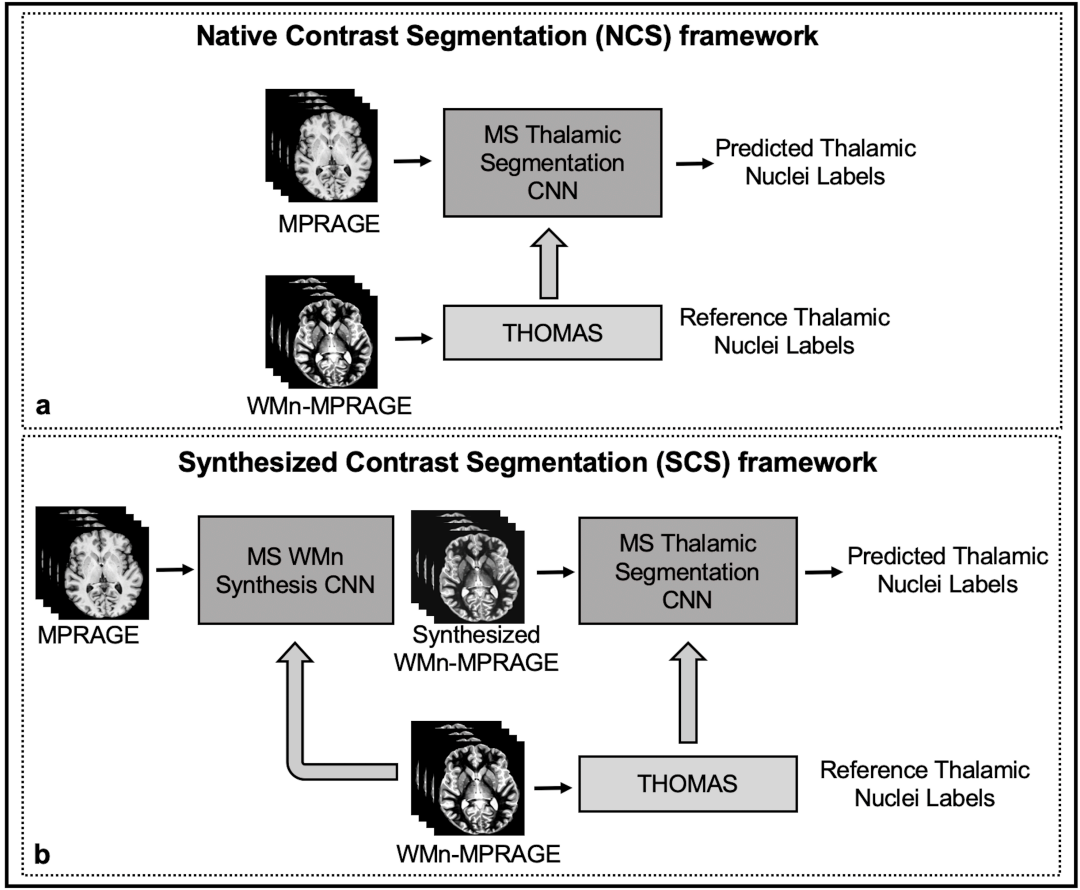

**Fig. 1**: Overview of the proposed multi-scale thalamic nuclei segmentation frameworks. (a) The thalamus and its individual nuclei are predicted directly on conventional MPRAGE images using the Native Contrast Segmentation (NCS) framework. (b) A multi-scale white-matter-nulled MPRAGE (WMn-MPRAGE) synthesis CNN is introduced prior to the segmentation CNN for synthesized contrast segmentation (SCS) framework. The ground truth labels are generated on the WMn-MPRAGE images using a multi-atlas based label fusion algorithm (THOMAS).

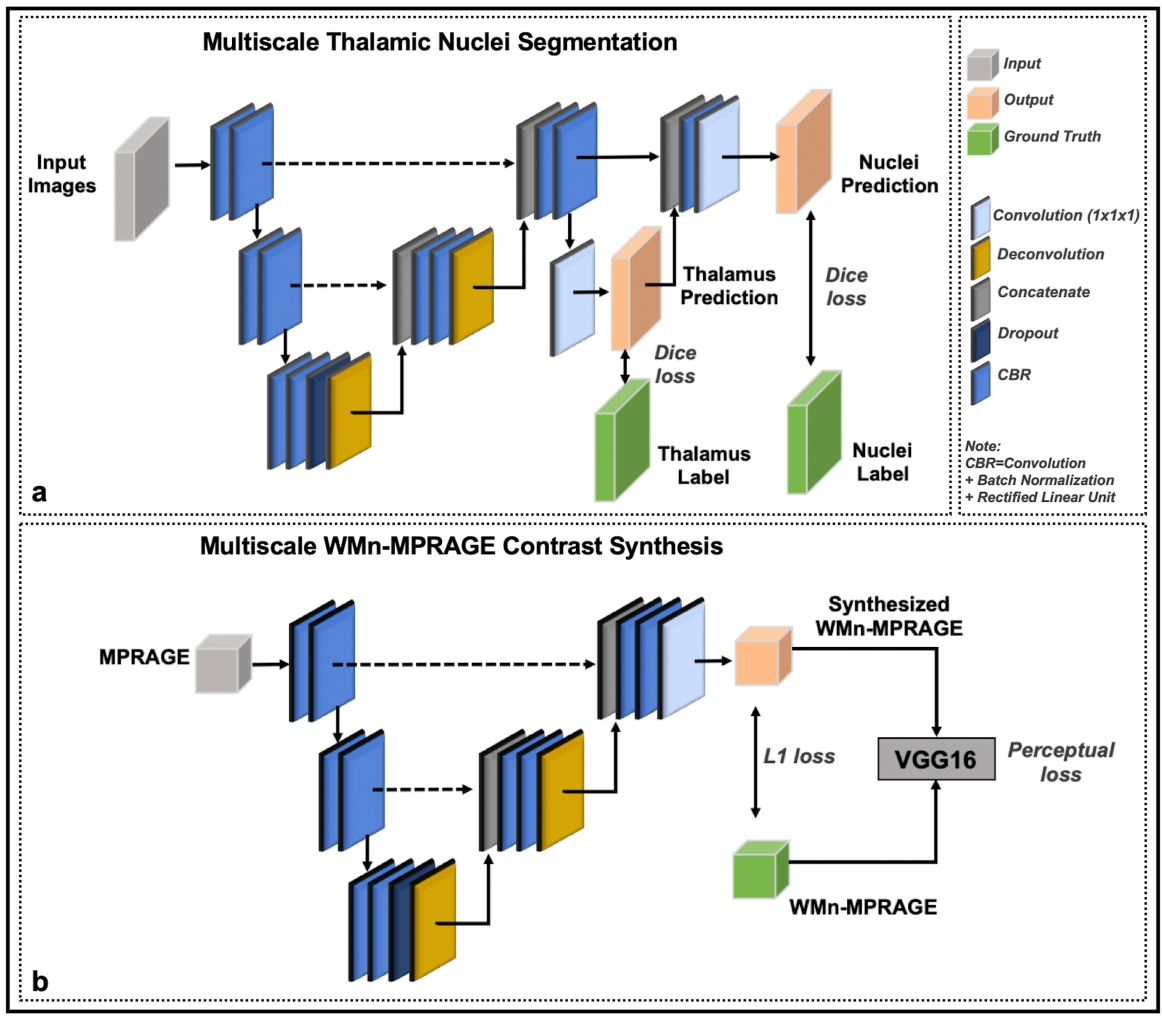

**Fig. 2**: Convolutional Neural Network (CNN) architectures for synthesis and segmentation. (a) A 3D multi-scale thalamic nuclei segmentation CNN and (b) a 3D multi-scale white-matter-nulled contrast synthesis CNN are shown. Both architectures have an encoder-decoder like framework with skip connections.

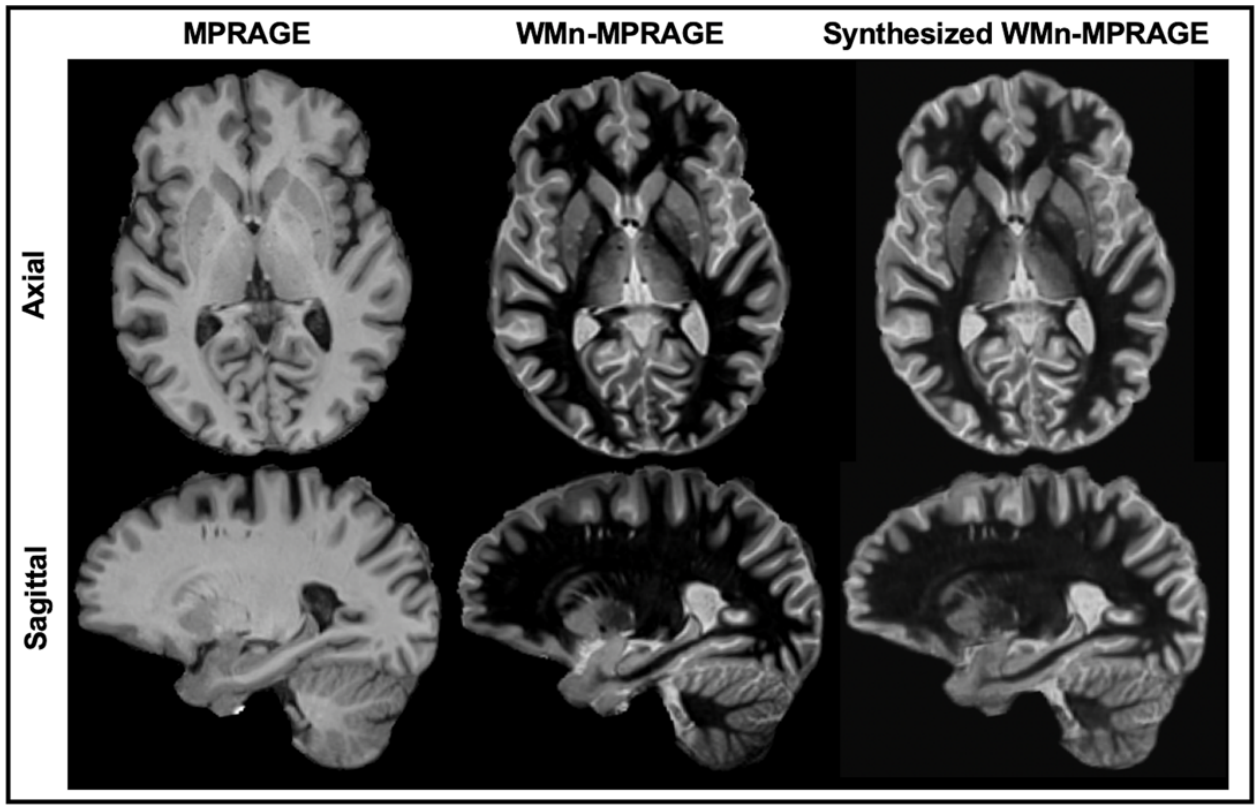

**Fig. 3**: White-matter-nulled MPRAGE (WMn-MPRAGE) synthesis. Representative axial and sagittal cross-sections of a synthesized image from a test subject are shown. The target WMn-MPRAGE images as well as input conventional MPRAGE images are also shown for reference. The synthesized WMn-MPRAGE images are able to mimic the contrast between thalamus and the surrounding white matter similar to the WMn-MPRAGE images

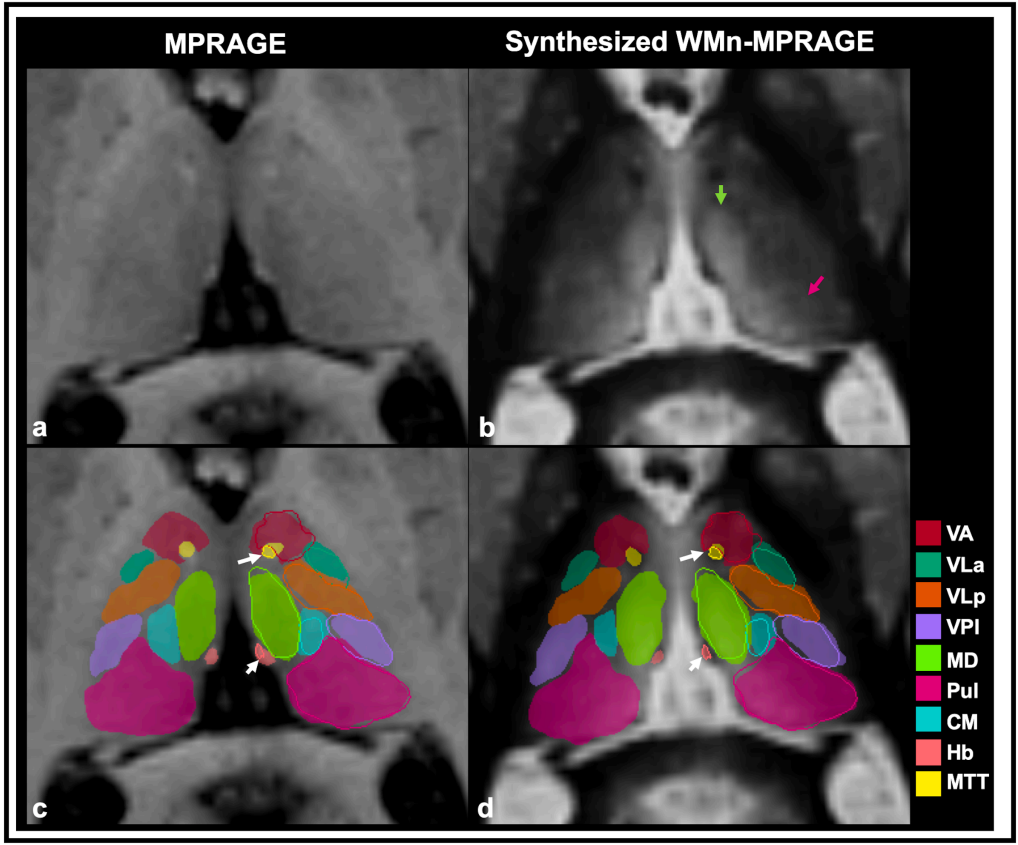

**Fig. 4**: Thalamic nuclei prediction on cerebrospinal fluid nulled MPRAGE and synthesized white-matter-nulled MPRAGE images. Top row (a-b): Representative axial cross-sections from a test subject for the different image contrast are shown. It is to be noted that synthesized WMn-MPRAGE contrast shows improved inter-thalamic contrast. The arrows in (b) highlight nuclei boundaries on synthesized WMn-MPRAGE image for MD and Pul nuclei. Bottom row: Individual thalamic nuclei labels predictions from the (c) native contrast and (d) synthesized contrast segmentation are overlaid on the images with reference THOMAS labels outlined on the left thalamus. The synthesized contrast segmentation CNN is able to accurately delineate even smaller nuclei such as VA, Hb, and MTT.

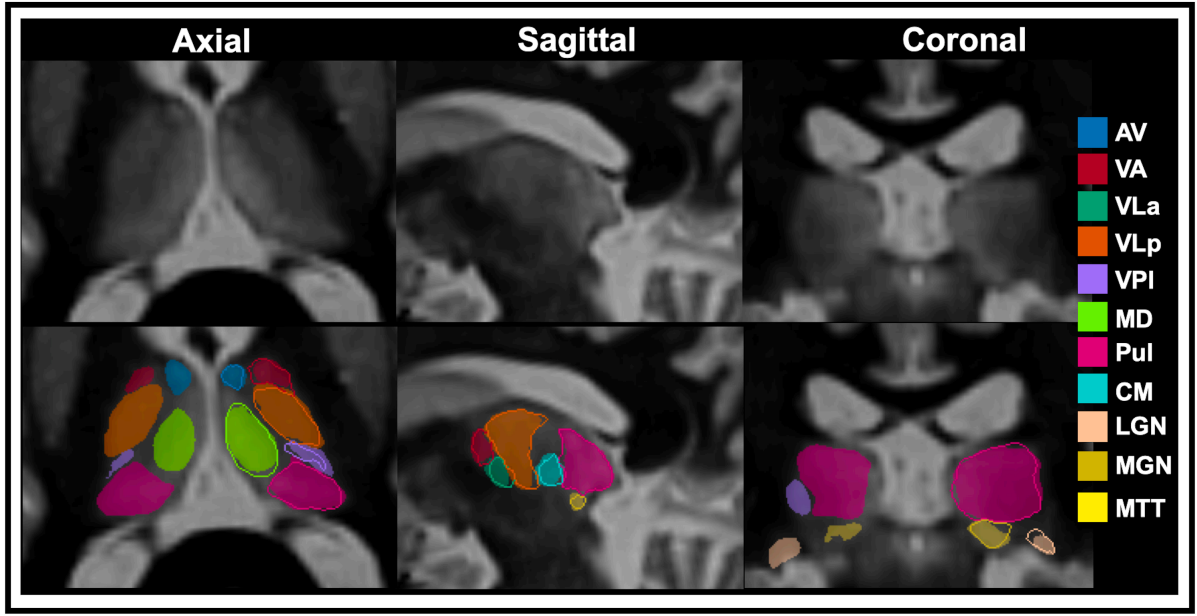

**Fig. 5**: Thalamic nuclei predictions from the synthesized contrast segmentation CNN. Multiplanar views of the test subject (top row) in Figure 4 with individual nuclei labels from the antral, ventral, posterior, and medial thalamic nuclei groups predicted using synthesized WMn-MPRAGE (bottom row) are shown. The corresponding ground truth labels from THOMAS are overlaid (color outlines) for the left thalamus.

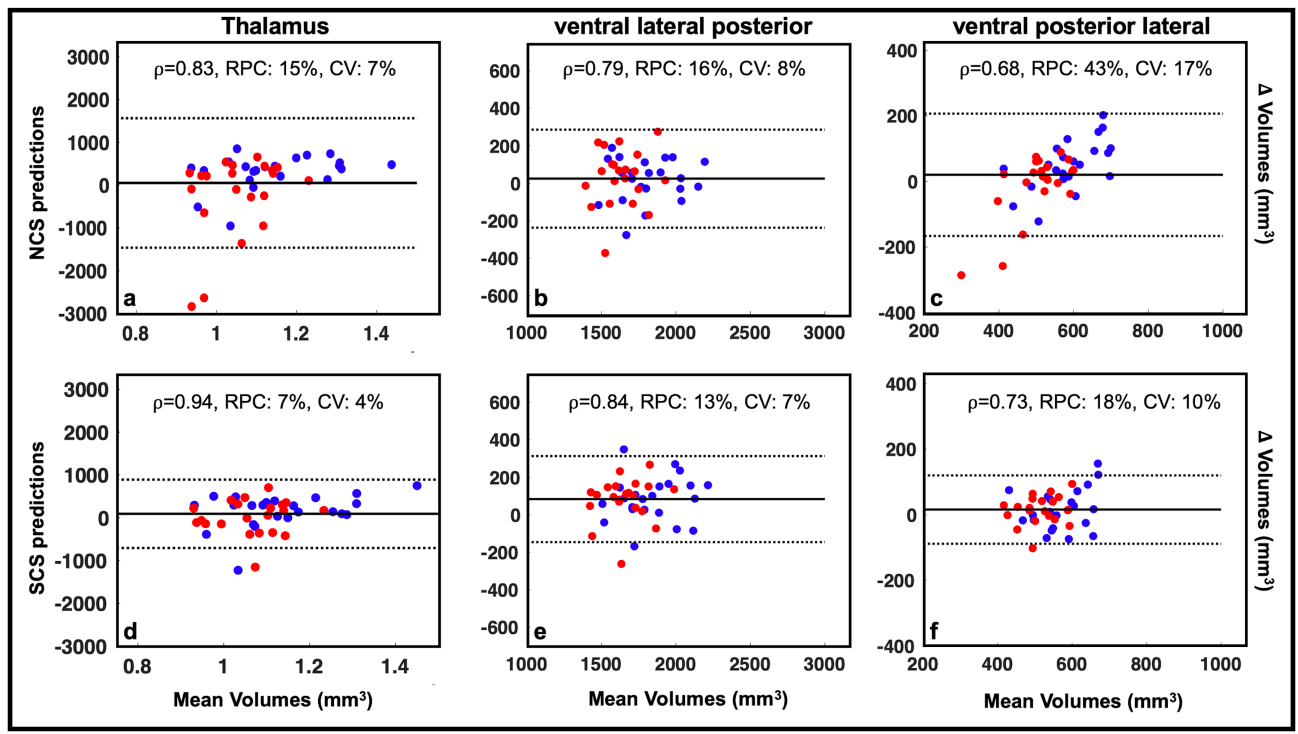

**Fig. 6**: Bland-Altman plots to assess agreement between predicted volumes and ground truth THOMAS labels. Bland-Altman plots for agreement in volumes for Thalamus (a), Ventral lateral posterior (b), and Ventral posterior lateral nucleus (c) for the segmentation CNN trained on conventional MPRAGE images (top row) are shown. The plots are also generated for the volumes predicted by the segmentation CNN trained on synthesized WMn-MPRAGE images (bottom row). The subjects are color-coded with blue for controls and red for subjects with alcohol use disorder. For each case, Pearson's correlation coefficient ($\rho$), percent repeatability coefficient (RPC), and coefficient of variation (CV) are also reported. It should be noted that the volumes from synthesized WMn-MPRAGE images have tighter limits of agreement.

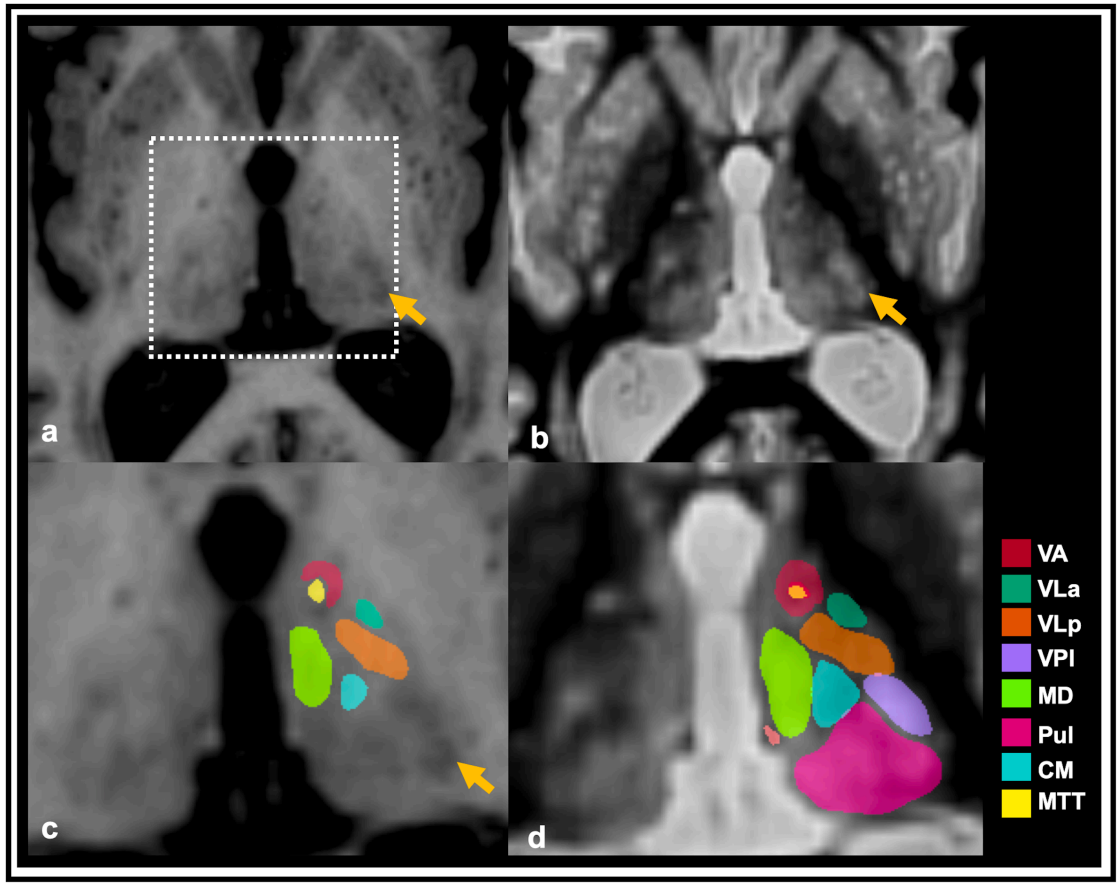

**Fig. 7**: Thalamic nuclei prediction on a subject with lesions in the thalamus. Axial cross-sections of the subject, with alcohol use disorder, are shown in the top row (a-b). The thalamic lesions (yellow arrows) appear bright in the WM-nulled contrast. The predictions from the corresponding NCS CNN (c) and SCS CNN (d) are shown. On this particular case, the segmentation using native MPRAGE contrast (NCS) fails to correctly delineate nuclei surrounding the lesions, especially Centromedian (CM), Pulvinar (Pul), and Ventral posterior lateral (VPl) nuclei.